\documentclass[runningheads]{llncs}
\usepackage{graphicx}
\usepackage{booktabs}
\usepackage{siunitx}
\usepackage{tabularx}
\usepackage{comment}
\usepackage{svg}
\usepackage{cite}

\begin{document}
\title{Bayesian Time-Series Classifier for Decoding Simple Visual Stimuli from Intracranial Neural Activity}
\titlerunning{Bayesian Time-Series Classifier}
\author{Navid Ziaei\inst{1}\orcidID{0000-0001-5834-6248} \and
Reza Saadatifard\inst{2} \and
Ali Yousefi\inst{2} \and
Behzad Nazari\inst{1} \and
Sydney S. Cash\inst{3} \and
Angelique C. Paulk\inst{3}\orcidID{0000-0002-4413-3417}}
\authorrunning{N. Ziaei et al.}
\institute{Department of Electrical and Computer Engineering, Isfahan University of Technology (IUT), Isfahan, Iran. \and
Department of Computer Science, Worcester Polytechnic Institute (WPI), Worcester, MA 01609, USA \and
Department of Neurology, Massachusetts General Hospital and Harvard Medical School, Boston, MA 02114, USA\\
\email{APAULK@mgh.harvard.edu}}
\maketitle              % typeset the header of the contribution
\begin{abstract}
Understanding how external stimuli are encoded in distributed neural activity is of significant interest in clinical and basic neuroscience. To address this need, it is essential to develop analytical tools capable of handling limited data and the intrinsic stochasticity present in neural data. In this study, we propose a straightforward Bayesian time series classifier (BTsC) model that tackles these challenges whilst maintaining a high level of interpretability. We demonstrate the classification capabilities of this approach by utilizing neural data to decode colors in a visual task. The model exhibits consistent and reliable average performance of 75.55\% on 4 patients’ dataset, improving upon state-of-the-art machine learning techniques by about 3.0 percent. In addition to its high classification accuracy, the proposed BTsC model provides interpretable results, making the technique a valuable tool to study neural activity in various tasks and categories. The proposed solution can be applied to neural data recorded in various tasks, where there is a need for interpretable results and accurate classification accuracy.

\keywords{Bayesian analysis  \and Neural decoding \and Interpretable modeling.}
\end{abstract}
\section{Introduction}
Neuroscientists have long sought methods to decode neural activity in hopes of restoring movement and communication for individuals with neurological injuries \cite{2:shanechi2019brain}, including stroke, spinal cord injury, brain trauma, and neurodegenerative diseases (e.g., Amyotrophic Lateral Sclerosis, ALS) using brain-computer interfaces (BCIs). Significant progress has been made in motor control \cite{3:kohler2017closed}, with advances also seen in the realms of speech \cite{4:anumanchipalli2019speech}, mood, and decoding neural activity corresponding to visual stimuli \cite{5:kay2008identifying,6:10.3389/fnsys.2016.00081,7:kosmyna2018attending}. Despite these discoveries, BCIs remain primarily in research endeavors, facing hurdles in terms of costs, risks, and technological challenges \cite{8:wolpaw2020brain}. The critical components of a BCI system involve feature extraction and accurate classification of neural activity related to different tasks or sensory input. However, several challenges exist in achieving highly accurate classifiers, including selecting the most informative and well-reasoned neural features and developing an interpretable classifier capable of utilizing limited datasets \cite{9:bashashati2007survey}. An interpretable classifier elucidates key neural signal features, like specific frequency bands or time periods, crucial for task performance. This insight enhances our understanding of the neural mechanisms.
Addressing these challenges is further complicated by the prohibitively large number of features used to decode neural activity. Features might encompass raw neural data, which include the measured voltage from single or multiple electrode contacts in non-invasive techniques such as electroencephalography (EEG), and invasive methods like intracranial EEG (iEEG) or blood-oxygen-level-dependent (BOLD) signal in functional MRI. Many researchers have suggested a variety of features derived from these neural recordings. These could range from simple statistical metrics of the signal in time or frequency domain, like mean, median, standard deviations, kurtosis, and skewness to more sophisticated features such as Common Spatial Patterns (CSPs) \cite{10:falzon2012analytic}, Higher-Order Crossing (HOC) features \cite{11:petrantonakis2009emotion}, Hjorth features \cite{12:oh2014novel}, and Auto-Regressive (AR) coefficients \cite{13:subha2010eeg}. As the EEG signal is non-stationary \cite{14:ubeyli2009statistics}, time-frequency domain methods such as wavelet transform (WT) have been used for feature extraction as well \cite{15:prochazka2008wavelet}. In addition to the types of features used, there can be multiple streams of the same data represented by different electrode channels. Many researchers employ data reduction techniques to handle this redundancy of information, such as Principal Component Analysis (PCA), Independent Component Analysis (ICA), or linear discriminant analysis (LDA) \cite{16:subasi2010eeg}. Another approach in data analysis focuses on selecting the most informative channels relative to the target classification task. This can be achieved through channel selection techniques, which can be categorized as either classifier-dependent (including wrapper and embedded-based methods) or classifier-independent (such as filter-based methods) \cite{17:alotaiby2015review}. In summary, different neural activity features can yield varied inferences, not always enhancing the understanding of encoding mechanisms. Thus, refining feature extraction and selection are essential for deriving relevant information that deepens our comprehension of brain processes.
Once features are chosen, classification commences, with considerable recent work involving deep-learning-based classifier models. While deep learning frameworks such as Convolutional Neural Networks \cite{18:gu2018recent} and Recurrent Neural Networks \cite{19:roy2019chrononet} have been applied to the decoding and classification of EEG signals to identify stimuli \cite{21:craik2019deep}, these methods might not be as applicable to invasive neural recordings like iEEG. This is primarily due to the limited availability of iEEG data compared to non-invasive EEG data. The drawbacks of these models include, 1) lack of interpretability, which corresponds to difficulty in identifying the necessary features for classification in many deep learning models, and 2) the requirement of vast amounts of data in the training phase of these models, which may not always be possible to collect when dealing with invasive recordings like iEEG \cite{22:roy2019deep}. Consequently, the limited data in invasive neural recordings necessitates the exploration of alternative methods that can effectively handle such constraints while maintaining accurate classification and interpretability.
In this study, we propose the Bayesian Time-Series Classifier (BTsC) model, designed to address the above-mentioned challenges. The BTsC allows us to identify the minimum number of channels necessary for stimulus classification from neural data. Furthermore, it enables the selection of neural features from different electrodes that provide optimal classification power, as well as the determination of the precise time period following a stimulus required for accurate stimulus classification. The proposed model can be trained effectively with limited data by leveraging the dynamics of local field potential (LFP) signals in two frequency subbands. Our proposed BTsC model employs a wrapper-based technique and greedy search in selecting channels and features for optimal classification. The pipeline of feature selection and classification used in the model can be applied to other classifiers, including Support Vector Machine (SVM) \cite{23:smola2004tutorial}, Long Short-Term Memory (LSTM) \cite{24:hochreiter1997long}, Naïve Bayes \cite{25:rish2001empirical}, etc. We applied this model to decode the presence of one or the other visual stimulus using LFP neural activity from multiple neural nodes’ activity, where our model shows high accuracy in classifying simple stimuli. In the following, we first outline our method of feature extraction from LFP signals and the development of the BTsC for stimulus decoding. Next, we assess the BTsC model's performance and compare it with other machine learning models. Lastly, we discuss our findings' implications for brain information processing and the potential applications of our model.

\section{Material and Methods}
\subsection{Dataset and Behavioral Task}
\label{sec:dataset-section}
\begin{figure}[h!]
\includegraphics[width=\textwidth]{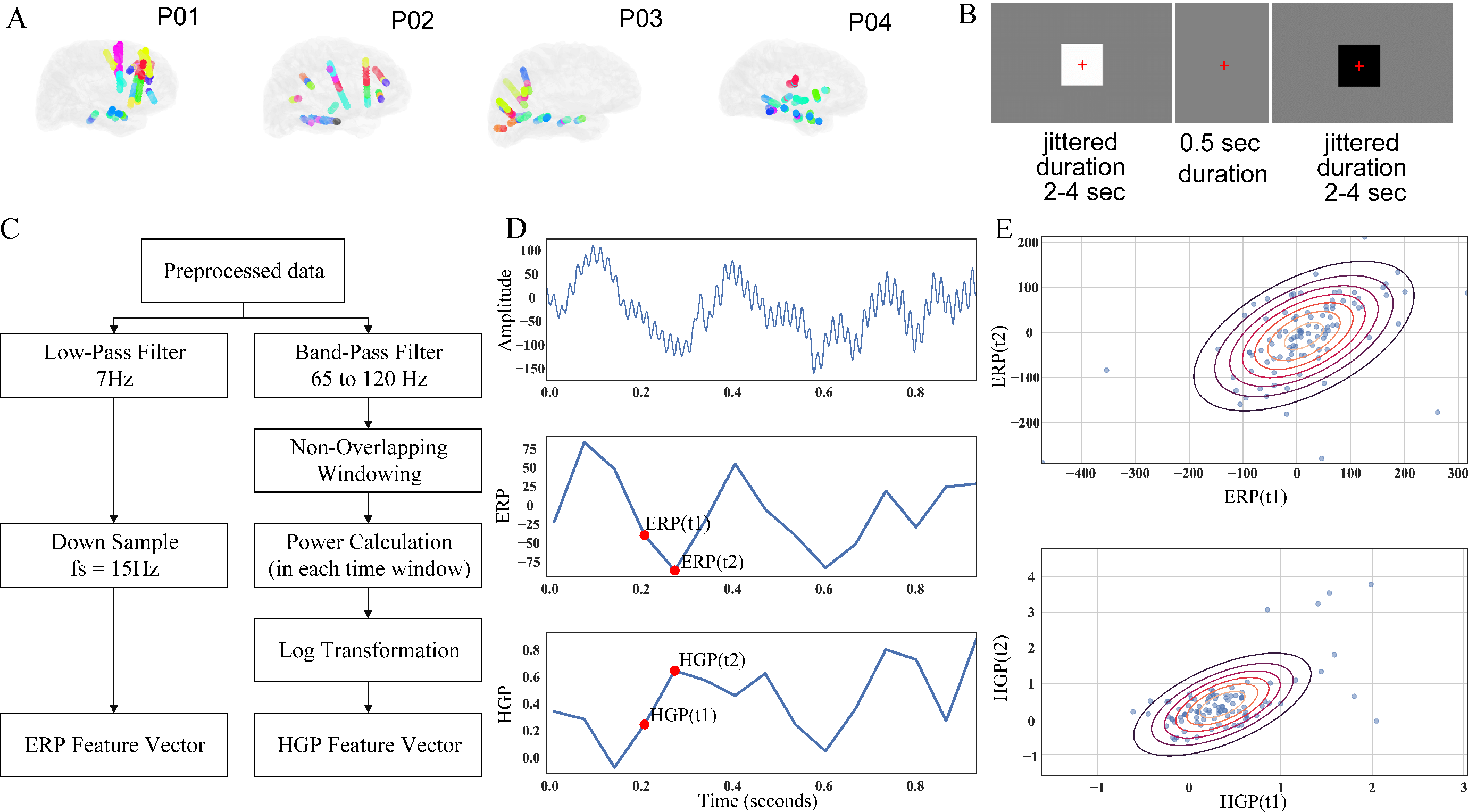}
\caption{Data Overview: (A) shows the electrode placement in four participants, (B) illustrates the Flicker Task paradigm performed in 100 trials, (C) outlines the steps for feature extraction, (D) displays the preprocessed single trial signal (top), ERP features (middle), and HGP features (bottom) extracted from the LTP02-LTP03 electrode for patient P04, and (E) presents a scatter plot of ERP (top) and HGP (bottom) features at times t1 and t2 for all trials recorded from the same electrode, indicating a Gaussian distribution.} \label{fig1}
\end{figure}

Four participants performed the visual stimulus task, while LFP was recorded from 278 sites via intracranial stereo EEG (Fig.~\ref{fig1}-A). Intracranial EEG recordings were made over the course of clinical monitoring for spontaneous seizures. Participants were implanted with multi-lead depth electrodes (also known as stereotactic EEG, sEEG) \cite{26:dykstra2012individualized}. All patients voluntarily participated after fully informed consent according to NIH guidelines, as monitored by the Massachusetts General Brigham (formerly Partners) Institutional Review Board (IRB). Participants were informed that participation in the tests would not alter their clinical treatment and that they could withdraw at any time without jeopardizing their clinical care. 
The Flicker task consisted of 100 trials where participants were presented with a red fixation cross for 0.5 seconds on a grey background (Fig.~\ref{fig1}-B). This was followed by the appearance of a single black or white square on the same grey background. The duration of the square varied randomly between 2 to 4 seconds. The color of the square for each trial was randomly selected from a sequence of black or white with equal chance. Participants were instructed to focus on the red fixation cross and count the number of black or white squares presented to enhance engagement. 

\subsection{Data Preprocessing: Intracranial LFP Data}
\label{sec:preprocessing-section}
Data analysis was performed using custom analysis code in MATLAB and Fieldtrip, an open-source software implemented in MATLAB\cite{27:oostenveld2011fieldtrip}. All data were subsequently decimated to 1000Hz, demeaned relative to the entire recording, and line noise and its harmonics up to 200Hz were removed by subtracting the band-passed filtered signal from the raw signal. Channels with excessive line noise or which had no discernible signal were removed from the analysis. In addition, we removed pathological channels with interictal epileptiform discharges (IEDs) using an automatic IED detection algorithm \cite{28:janca2015detection} (version v21, default settings except -h at 60; http://isarg.fel.cvut.cz). We removed channels that had detected IEDs greater than 6.5 IEDs/minute The remaining electrodes were subsequently bipolar re-referenced relative to nearest neighbors to account for volume conduction.

\subsection{Extracting Neural Features for Decoding}
\label{sec:feature-section}
We focus on two categories of features known to encode stimulus information, namely the low-frequency event-related potentials (ERPs) and the high gamma power (HGP) following image onset (see Fig.~\ref{fig1}-C).
For the ERP neural features, we filter the LFP in the theta (3-7 Hz) and delta (0-3 Hz) frequency bands \cite{29:vidal2010category} using a low-pass filter with a cut-off frequency of 7 Hz. By applying the low-pass filter, we conform to the Nyquist-Shannon sampling theorem, which allows us to down-sample the filtered data to 15 Hz without losing crucial information. This results in 15 features (each feature being an individual time step) for ERP signals per electrode. Each sample of the ERP feature vector includes a weighted average of multiple samples of the original time series data in a specific time interval. Under the central limit theorem assumption, we can assume that each element of these vectors follows a normal distribution.
As HGP has also been shown to encode visual stimuli \cite{29:vidal2010category,30:liu2009timing}, we band-pass filter the LFP from 65 to 120 Hz. Power is then calculated in 67-millisecond nonoverlapping windows after stimulus onset, generating the same number of features per each channel as the ERP. A subsequent step involves using a log transformation. This transformation compresses the range of high values and expands the range of low values, resulting in a distribution that is more symmetric and closer to a normal distribution. This justifies its use as an input feature vector to our model \cite{31:box1964p}.
In the procedure described, each LFP recording channel results in two feature vectors, one for ERP and one for HGP, each represented as a time series. We model these vectors as multivariate normal distributions for use in the BTsC model, which we'll discuss further in the next section.

\subsection{Bayesian Time-series Classifier (BTsC)}
\label{sec:model-section}
In section \ref{sec:feature-section}, we described how ERP and HGP feature vectors are acquired from each LFP recording channel. To simplify the description, we assume that each channel is associated with a single feature vector, and we refer to the classifier trained on this feature vector as a single-channel classifier. This simplification does not compromise the broader applicability of the model. To present the BTsC model, we start by detailing the single-channel classifier's construction and the process of determining optimal time periods. We then discuss combining single-channel classifiers into a multi-channel classifier and selecting the minimum subset of classifiers needed for maximum accuracy. The BTsC aims to determine the fewest feature vectors necessary for effective stimulus classification.
Single-channel classifier. Let us assume that the pre-processed neural recording for the $c^{th}$ electrode is represented by the feature vector defined by $\mathbf{x}^{c}=\{\mathbf{x}_{1}^{c},\mathbf{x}_{2}^{c},…,\mathbf{x}_{d}^{c}\}$, where $d$ is the length of observation period, and ${\mathbf{x}_i}^{c}$ is the $i^{th}$ sample of the feature vector for the $i^{th}$ time interval after the stimulus onset. As discussed in section \ref{sec:feature-section}, we assume that each element of $\mathbf{x}^{c}$  follows a normal distribution. We further assume the joint distribution of $\mathbf{x}^{c}$ follows a multivariate normal, where the dependency across time points allows us to characterize the temporal dynamics of observed neural features. For the model, we build the conditional distribution of $\mathbf{x}^c$ given stimulus, $I\in\{0,…,k \}\equiv\{(stimulus 1, ..., stimulus K)\}$. The conditional distribution of $\mathbf{x}^c$ is defined by:
\begin{equation}
\mathbf{x}^{c} |I\sim\mathcal{N}(\mathbf{\mu}_{I}^{c},\mathbf{\Sigma}_{I}^{c})
\end{equation}
\begin{equation}
p(\mathbf{X}= \mathbf{x}^{c}|I)=\frac{1}{(2\pi)^{d/2}|\mathbf{\Sigma}_{I}^{c}|^{1/2}}
\exp\left(-\frac{1}{2}(\mathbf{x}^{c}-\mathbf{\mu}_{I}^{c})^T {\mathbf{\Sigma}_{I}^{c}}^{-1}(\mathbf{x}^{c}-\mathbf{\mu}_{I}^{c})\right)
\end{equation}
where $\mathbf{\mu}_{I}^{c}$ and $\mathbf{\Sigma}_{I}^{c}$ are the mean and covariance of $\mathbf{x}^{c}$ given stimuli $I$. Given $\mathbf{\mu}_{I}$ and $\mathbf{\Sigma}_{I}$, we can construct a Bayesian classifier to compare the posterior probabilities of different stimuli. The assigned class is the one with the highest posterior probability, defined by:
\begin{equation}
    \forall j\neq k: L(I=k \mid \mathbf{X}=\mathbf{x}^{c})\geq L(I=j \mid \mathbf{X}= \mathbf{x}^{c})
\end{equation}
$L\left(I\mid \mathbf{X}=\mathbf{x}^{c}\right)$ corresponds to the posterior distribution of stimulus $I$, given the observed neural features, defined by:
\begin{equation}
    L\left(I \mid \mathbf{X}= \mathbf{x}^{c}\right) \propto p(\mathbf{X}= \mathbf{x}^c \mid I)p(I)
\end{equation}
where $p(I)$ is the stimulus prior probability. To build our single-channel classifier, we require only the mean and covariance of each neural recording ($\mathbf{x}^{c}$) per stimulus. Using the training dataset $\mathcal{D}$, we find the mean and covariance matrix for each time series. To obtain a robust estimation of the covariance matrix, the number of required samples must be on the order of $d^{2}$. Estimating the covariance matrix can result in a biased estimation given the limited training data \cite{32:ledoit2004well}. Our solution is to find the minimum length of the observation ($d_{minimal}^{c}$) starting from the onset, providing the highest accuracy with the cross-validated result. Using this approach, we can address the limited training dataset in our estimation of the covariance matrix. In the case of the multivariate normal distribution, we can obtain the marginal distribution of a subset of the neural data; any marginalized distribution remains a multivariate normal. With this in mind, we can examine the posterior of each class of data as a time-dependent function, identifying the stimulus from a subset of neural data $\mathbf{x}^{c}$. We denote this posterior as $L_{j}$, signifying a marginalized version of the overall posterior distribution $L$, but only considering the first $j$ features $\{\mathbf{x}_{1}^{c},\mathbf{x}_{2}^{c},…,\mathbf{x}_{j}^{c}\}$. We introduce the concept of $C_{j} (\mathcal{D})$, representing the cross-validated classification accuracy of our model on the dataset $\mathcal{D}$, using a marginalized posterior distribution with the first $j$ features. For each classifier, the minimal time, denoted as $d_{minimal}^{c}$, is defined as follows:
\begin{equation}
    d_{\text{minimal}}^{c} = \arg\max_{j}C_j(\mathcal{\mathcal{D}}) 
\end{equation}
This suggests that $d_{minimal}^{c}$ represents the smallest set of features necessary to optimize our model's performance, in accordance with the constraints of the k-fold cross-validation method being used.

\subsubsection{Multi-channel Classifier.} We construct our BTsC model initially based on a single-channel classifier, which turns to Quadratic Discriminant Analysis (QDA) for time series \cite{33:srivastava2007bayesian}. In practice, we have multiple channels, with each channel having two feature vectors. Classifiers trained on these feature vectors can be combined to achieve higher classification accuracy. Equation (4) defines the classifier model for the $c^{th}$ feature vector. We found that the single-channel classifier accuracy is limited, and in the best case, it is about or less than $75\%$. To attain higher classification accuracy, we expand our single-channel QDA classifier to account for multiple channels. We employ two solutions to adapt the classifier for multi-channel neural recordings, resulting in an ensemble-based classifier that enhances accuracy and robustness. The first solution expands the model directly to multiple channels. The second solution is based on the majority voting rule of $C$ different classifiers, where $C$ is the number of possible channels. 

\paragraph{Multi-channel Likelihood Method.} For a multi-channel classifier, we assume that different feature vectors are conditionally independent given the stimulus. The joint conditional distribution of all recordings is defined by:
\begin{equation}
    p(\mathbf{X}_{1}=\mathbf{x}^{1},\ldots,\mathbf{X}_{C}=\mathbf{x}^{C} |I)=p(\mathbf{X}_1=\mathbf{x}^1 |I) \ldots p(\mathbf{X}_{C}=\mathbf{x}^C |I)
\end{equation}
Where $I$ represents the stimulus. We can construct each single-channel model similar to the one defined in Equation (2). The posterior distribution for each multi-channel neural recording is defined by:
\begin{equation}
    L(I;\mathbf{X}_1=\mathbf{x}^1,\ldots,\mathbf{X}_C=\mathbf{x}^C ) \propto p(\mathbf{X}_1=\mathbf{x}^1 |I) \ldots p(\mathbf{X}_C=\mathbf{x}^C |I)p(I)
\end{equation}

Utilizing all channels in the model is not practical as some may lack informative features for classification. Also, multiplying likelihoods from all channels can complicate computation due to smaller values. Hence, identifying the most informative channels through channel subset selection is necessary for accurate classification. We will discuss this challenge in the classification subset selection section.

\paragraph{Maximum Voting Method.} The outcomes of single-channel classifiers can be combined using the voting method to achieve higher classification accuracy. In a poll of $C$ single-channel classifiers, each classifier contributes a vote towards a class label based on its independent prediction. The class which receives the most votes, representing the majority opinion among the classifiers, is selected as the final output class \cite{33:srivastava2007bayesian}. 

\subsubsection{Classifiers Subset Selection.} We can combine single-channel classifiers to construct a multi-channel classifier using either of these two methods. The process of selecting the optimal subset of feature vectors is based on an adaptive (or greedy) search. It begins with a single channel with the best performance using k-fold cross-validation and then examines which other channel can be added to it. All possible combinations are evaluated, and the one with the best performance is selected. Through this adaptive search, the minimal number of channels that reach the highest classification accuracy with the highest confidence can be determined with cross-validation.

\section{Results}
\begin{figure}[htbp!]
     \centering
     \includegraphics[width=\textwidth]{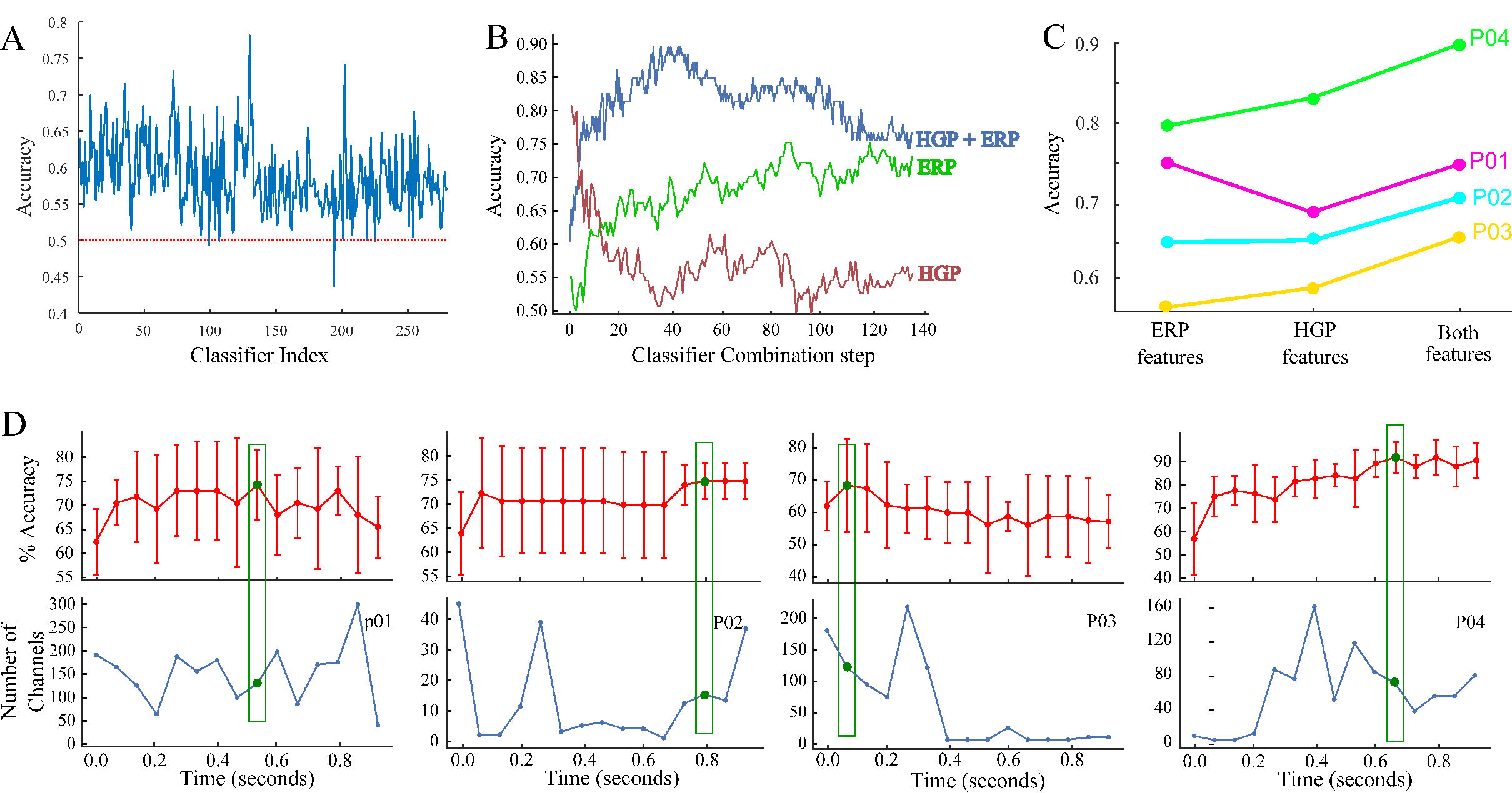}
     \caption{Results: (A) illustrates the performance of individual classifiers. (B) shows accuracy evolution during channel combination steps for participant P05. (C) shows the comparison between the BTsC performance using ERP, HGP, and both ERP and HGP features to assess the impact of neural features. (D) displays the accuracy of the model at different time points for participants P01 to P05.}
     \label{fig:result}
 \end{figure}

The BTsC model can be applied to different mental tasks with various features. Here, we investigated the application of BTsC in the visual task described in section \ref{sec:dataset-section}. We trained the BTsC using low-frequency (ERP, $<7 Hz$) and high-frequency power (HGP; 65-120 Hz) dynamics following image onset to test if we can decode visual stimuli from across the brain at a high temporal resolution (67 ms). Then, we identified the features and time points for maximum decoding accuracy (Fig.~\ref{fig:result}). Furthermore, to validate the results obtained with the BTsC model, we compared them with the decoding outcomes of seven additional machine learning (ML) algorithms on visual stimuli.
Optimal Stimulus Decoding Time and Features in the Model
Following the model subset selection, we discovered that the channels and features that survived the selection criteria and contributed to the BTsC models were from multiple electrodes across multiple brain regions. BTsC enabled us to evaluate the performance of each feature vector, ERP and HGP, on individual channels and to determine the optimal timing post-image onset for superior performance. From this analysis, we discerned which regions and which features exhibited the fastest responses. Additionally, we found that combining these feature vectors leads to a boost in performance (Fig.~\ref{fig:result}-B). Upon analyzing the BTsC results for ERP, HGP, and their combined utilization, we discovered that leveraging both ERP and HGP features enhances the decoding model's accuracy in the visual task (Fig.~\ref{fig:result}-C).
After identifying the time window after image onset with the peak accuracy for the single time or cumulative BTsC models, we found that the time of maximum accuracy after image onset was below 0.8 seconds. The accuracy at each time point and the number of utilized feature vectors are depicted in Fig.~\ref{fig:result}-D.
\subsection{Machine Learning Decoding}
To test if decoding results, which support the distributed information flow hypothesis \cite{34:sabesan2009information} are particular to the BTsC model, we applied seven machine learning models to the same neural features (ERP and HGP) and participant data. The machine learning classifiers include SVM \cite{23:smola2004tutorial}, Logistic Regression (LR) \cite{35:james2013introduction}, Naïve Bayes (NB) \cite{25:rish2001empirical}, Random Forest (RF) \cite{35:james2013introduction}, Multi-Layer Perceptron (MLP) \cite{36:gardner1998artificial}, LSTM \cite{24:hochreiter1997long} and EEGNet \cite{37:lawhern2018eegnet}.
\begin{comment}
\begin{table*}[h!]
\centering
\caption{Peformance Comparison across Different ML Techniques: The mean and standard deviation of 5-fold cross-validation f-measure  for all models.}
\label{tab:modelPerformance}
\begin{tabular*}{\textwidth}{@{\extracolsep{\fill}}lcccc}
\toprule
\textbf{Model} & \textbf{P01} & \textbf{P02} & \textbf{P03} & \textbf{P04} \\
\midrule
BTsC (likelihood) & $74.61 \pm 5$ & $73.75 \pm 4$ & $66.51 \pm 7$ & $85.41 \pm 7$ \\
BTsC (voting) & $75.11 \pm 10$ & $70.61 \pm 6$ & $66.49 \pm 10$ & $90.00 \pm 6$ \\
EEGNet & $72.04 \pm 7$ & $68.11 \pm 7$ & $67.42 \pm 9$ & $85.32 \pm 10$ \\
SVM & $61.25 \pm 13$ & $70.00 \pm 14$ & $58.55 \pm 9$ & $81.21 \pm 7$ \\
Naïve Bayes & $70.00 \pm 6$ & $68.75 \pm 8$ & $51.18 \pm 11$ & $74.84 \pm 16$ \\
Random Forest & $60.00 \pm 6$ & $62.50 \pm 3$ & $55.28 \pm 4$ & $72.93 \pm 6$ \\
MLP & $67.81 \pm 9$ & $67.71 \pm 11$ & $58.55 \pm 10$ & $71.97 \pm 8$ \\
Logistic Regression & $66.25 \pm 11$ & $67.50 \pm 7$ & $53.12 \pm 5$ & $75.05 \pm 12$ \\
LSTM & $64.62 \pm 8$ & $67.71 \pm 7$ & $52.77 \pm 9$ & $70.05 \pm 7$ \\
\bottomrule
\end{tabular*}
\end{table*}
\end{comment}
\begin{table*}[htbp!]
\centering
\caption{Performance Comparison across Different ML Techniques: The table depicts the mean and standard deviations (presented in parentheses) of 5-fold cross-validation accuracy for all models, including BTsC (L) and BTsC (V), where (L) and (V) represent combination methods using likelihood and voting, respectively.}
\label{tab:modelPerformance}
\begin{tabular*}{\textwidth}{@{\extracolsep{\fill}} l 
  S[table-format=2.1(2)]S[table-format=2.1(2)]
  S[table-format=2.1(2)]S[table-format=2.1(2)]
  S[table-format=2.1(2)]S[table-format=2.1(2)]
  S[table-format=2.1(2)]S[table-format=2.1(2)]}
\toprule
& \multicolumn{2}{c}{\textbf{P01}} 
& \multicolumn{2}{c}{\textbf{P02}}
& \multicolumn{2}{c}{\textbf{P03}}
& \multicolumn{2}{c}{\textbf{P04}} \\
\cmidrule(lr){2-3}
\cmidrule(lr){4-5}
\cmidrule(lr){6-7}
\cmidrule(lr){8-9}
\textbf{Model} & \multicolumn{1}{c}{Acc.} & \multicolumn{1}{c}{F1} 
& \multicolumn{1}{c}{Acc.} & \multicolumn{1}{c}{F1}
& \multicolumn{1}{c}{Acc.} & \multicolumn{1}{c}{F1}
& \multicolumn{1}{c}{Acc.} & \multicolumn{1}{c}{F1} \\
\midrule
BTsC (L)    & 74.6(5)   & \textbf{75.2(8)}   & \textbf{73.7(4)}   & \textbf{71.0(9)}   & 66.5(7)   & 62.0(8)   & 85.4(7)   & 83.1(3) \\
BTsC (V)    & \textbf{75.1(9)}  & 74.3(7)   & 70.6(6)   & 69.8(9)   & 66.4(10)  & 67.3(7)   & \textbf{90.0(6)}   & \textbf{89.1(4)} \\
EEGNet      & 72.0(7)   & 73.2(9)   & 68.1(7)   & 70.1(11)  & \textbf{67.4(9)}   & \textbf{68.1(9)}   & 85.3(10)  & 84.2(8) \\
SVM         & 61.2(13)  & 65.2(8)   & 70.0(14)  & 73.0(9)   & 58.5(9)   & 51.0(10)  & 81.2(7)   & 81.1(10) \\
NB          & 70.0(6)   & 69.4(8)   & 68.7(8)   & 70.7(8)   & 51.1(11)  & 54.0(18)  & 74.8(16)  & 70.9(9) \\
RF          & 60.0(6)   & 58.5(3)   & 62.5(3)   & 63.0(3)   & 55.2(4)   & 56.1(8)   & 72.9(6)   & 71.2(5) \\
MLP         & 67.8(9)   & 63.2(6)   & 67.7(11)  & 68.1(9)   & 58.5(10)  & 55.9(10)  & 71.9(8)   & 72.5(7) \\
LR          & 66.2(11)  & 66.0(8)   & 67.5(7)   & 66.8(13)  & 53.1(5)   & 46.5(5)   & 75.0(12)  & 76.1(10) \\
LSTM        & 64.6(8)   & 63.2(6)   & 67.7(7)   & 64.2(9)   & 52.7(9)   & 50.0(11)  & 70.0(7)   & 71.9(11) \\
\bottomrule
\end{tabular*}
\end{table*}

\section{Discussion}
Our investigation has yielded significant insights into the effectiveness of the BTsC in decoding visual stimuli from intracranial EEG recordings. The BTsC model's integration of the ERP and HGP features has demonstrated a remarkable capacity for classifying visual stimuli, outperforming other machine learning models including SVM, Logistic Regression, Naïve Bayes, Random Forest, MLP, LSTM, and EEGNet.
By leveraging the BTsC model, we achieved an average accuracy of $75.55\%$ in classifying stimuli (Table~\ref{tab:modelPerformance}). This is a noteworthy outcome, particularly given the inherent complexity of neural data, inter-subject variability in iEEG electrode position, and the challenges associated with decoding neural signals. Our results show that the BTsC model can effectively capture the distributed information flow within the brain, with its simplicity offering robustness against limited training data. In comparison, other methods face challenges such as overfitting, interpretability, and scalability issues. For example, complex models like EEGNet can result in overfitting due to their complex architecture, making them less reliable when dealing with limited data \cite{21:craik2019deep}. On the other hand, simpler models like Naïve Bayes rely on the assumption of features' independence, which is often unrealistic in real-world applications.
The key feature of our model is its interpretability, essential for validating predictions and understanding brain functions \cite{38:zimek2018there}. Our BTsC model offers a clear view of which areas of the brain are encoding stimuli and at what time these features exhibit the most discriminative power. Additionally, this model possesses flexibility in utilizing diverse feature vectors and can provide insights into the specific contributions of each vector to the final outcome. 
In our study, we leveraged both ERP and HGP dynamical features individually and in combination within the BTsC. While both ERPs and HGP independently provided significant information for decoding visual stimuli, we found that combining these features led to a marked increase in classification accuracy. This suggests that these neural features, while independently informative, offer complementary information that enhances the decoding power when combined, reflecting the complex and distributed nature of brain processing \cite{39:sporns2005human}. Thus, the integration of multiple neural features could contribute to more accurate and robust models for neuroscience applications.
Further expanding our research, we conducted an experiment to determine the most informative time period after image onset for training the BTsC model. This aspect is crucial, as it helps establish the optimal window for capturing the most discriminative neural features. In this experiment, we trained the BTsC model using different time windows post-image onset. Consequently, we determined the optimal time window post-image onset for training the BTsC model for each patient (Fig.~\ref{fig:result}-D). Moreover, this experiment revealed that in shorter time windows, HGP feature vectors are more involved than ERP. 
Despite the encouraging results, our model has limitations. The assumption of channels' independence is a significant limitation, which we intend to address in future iterations. We plan to refine our model to account for possible dependencies and correlations between features and electrodes, which could further enhance the predictive accuracy of the BTsC model.

\section{Conclusion}
In this study, we introduced a novel Bayesian Time-series Classifier that utilizes both low-frequency event-related potentials and high gamma power features to effectively decode visual stimuli from intracranial EEG recordings. The BTsC model identifies encoding brain areas, discriminative features, and optimal time windows, outperforming other classifiers by $3\%$ in accuracy due to its ability to capture distributed information flow. With its demonstrated success in decoding simple visual information and accommodating individual variations, this model holds promise for applications in neuroscience and clinical settings, including brain-computer interfaces and neural prosthetics. Future research will broaden the scope to include more cognitive tasks and modalities, personalize neurotechnology through additional neural features, and explore the impact of different covariance structures on our Bayesian Time-Series Classifier model.

\bibliographystyle{unsrt}
\bibliography{main}{}
%\end{thebibliography}
\end{document}